\documentclass{article}
\usepackage{spconf,amsmath,graphicx}

\usepackage{cite}
\usepackage{amsmath,amssymb,amsfonts}
\usepackage{algorithmic}
\usepackage{graphicx}
\usepackage{textcomp}
\usepackage{xcolor}
\usepackage{times}
\usepackage{latexsym}
\usepackage{soul}
\usepackage{url}
\usepackage{graphicx}
\usepackage{amsmath}
\usepackage{xcolor}
\usepackage[ruled]{algorithm2e}

\usepackage{stfloats}
\usepackage{booktabs}
\usepackage[small]{caption}

\title{L2RS: A Learning-to-Rescore Mechanism for Automatic Speech Recognition}

\name{\begin{tabular}{c}Yuanfeng Song$^{\star \dagger}$, Di Jiang$^{\dagger}$, Xuefang Zhao$^{\Diamond}$, Qian Xu$^{\dagger}$ \\
Raymond Chi-Wing Wong$^{\star}$, Lixin Fan$^{\dagger}$, Qiang Yang$^{\star \dagger}$ \end{tabular}}\vspace{-1mm}
\address{
  $^{\star}$Department of CSE, The Hong Kong University of Science and Technology, Hong Kong, China \\
  $^{\dagger}$AI Group, WeBank Co., Ltd, Shenzhen, China 
  $^{\Diamond}$Tsinghua University, Shenzhen, China 
}

\begin{document}

\maketitle

\begin{abstract}
Modern Automatic Speech Recognition (ASR) systems primarily rely on scores from an Acoustic Model (AM) and a Language Model (LM) to rescore the $N$-best lists.
With the abundance of recent natural language processing advances, the information utilized by current ASR for evaluating the linguistic and semantic legitimacy of the $N$-best hypotheses is rather limited.
In this paper, we propose a novel \emph{Learning-to-Rescore} (L2RS) mechanism, which is specialized for utilizing a wide range of textual information from the state-of-the-art NLP models and automatically deciding their weights to rescore the $N$-best lists for ASR systems.
Specifically, we incorporate features including BERT sentence embedding, topic vector, and perplexity scores produced by $n$-gram LM, topic modeling LM, BERT LM and RNNLM to train a rescoring model. We conduct extensive experiments based on a public dataset, and experimental results show that L2RS outperforms not only traditional rescoring methods but also its deep neural network counterparts by a substantial improvement of 20.67\% in terms of NDCG@10. L2RS paves the way for developing more effective rescoring models for ASR.
\end{abstract}
\begin{keywords}
automatic speech recognition, $N$-best list rescoring, feature engineering, learning-to-rescore
\end{keywords}
\section{Introduction}
\label{sec:intro}

Due to the ubiquitous existence of speech in daily life, Automatic Speech Recognition (ASR) is gaining significant momentum in recent years. However, current ASR systems primarily rely on scores produced by an Acoustic Model (AM) and a Language Model (LM) to rank the $N$-best lists, and usually the 1-best of the hypotheses is selected as the final recognition result. For computing the LM score, the back-off $n$-gram LM is prominently used for many years due to its simplicity and reliability \cite{bellegarda2004statistical}. However, $n$-gram LM is rather simplistic and heavily limited in its ability of modeling language context such as long-range dependencies.

In order to alleviate the above problem of $n$-gram LM, the mechanism of $N$-best list rescoring is proposed and proven to be effective to significantly improve the ASR performance \cite{jurafsky2000speech}. For example, Discriminative Language Model (DLM) is proposed in \cite{Roark:2007:DNL:1221595.1221969, 6064876, roark2004discriminative} and it utilizes features such as the ASR errors to train a discriminative model for $N$-best list rescoring. With the arise of deep neural network in ASR, RNNLMs \cite{mikolov2010recurrent,mikolov2011extensions} and LSTM-based LMs \cite{erdogan2016multi} are becoming popular models for $N$-best list rescoring. More recently, Ogawa et al. \cite{ogawa2018rescoring} propose a Encoder-Classifier Model (EC-Model) which trains a classifier to compare between the pairs in $N$-best lists to do the rescoring. With their merits, each of these methods utilizes quite limited information, and the vast arsenal of the state-of-the-art models for gauging linguistic and semantic legitimacy is heavily underused. For example, the common word embeddings (from Word2Vec \cite{mikolov2013distributed}, Speech2Vec \cite{chung2018speech2vec} to BERT \cite{devlin2019bert}) are hard to utilize under existing rescoring frameworks. 

In contrast to the conventional approach that simply adds up a LM score and an AM score for $N$-best list rescoring, we propose a novel \emph{Learning-to-Rescore} (L2RS) mechanism, which for the first time formalizes the $N$-best list rescoring as a learning problem. L2RS utilizes a wide range of features with automatic optimized weights to rank the $N$-best lists for ASR and selects the most promising one as the final decoding result. The efficacy of L2RS relies on the design of the features. We extract features using BERT sentence embedding, topic vector and perplexity scores given by probabilistic topic models such as LDA \cite{blei2003latent}, neural network based language models, such as RNNLM \cite{mikolov2010recurrent}, and BERT LM\cite{devlin2019bert}, together with the score given by an acoustic model. By combining all these features together, L2RS learns a rescoring model using RankSVM \cite{cao2006adapting} algorithm. 
Since each feature reflects one perspective from the linguistic and semantic legitimacy of the $N$-best hypotheses, L2RS achieves superior performance by ensembling the information from all these evaluation metrics. The main contribution of the paper is summarized as follows:

$\bullet$  To the best of our knowledge, this is the first work that formalizes the $N$-best list rescoring problem as a Learning-to-Score problem for ASR.

$\bullet$  We propose a novel L2RS framework dedicated for ASR, which can easily incorporate various state-of-the-art NLP models to extract features. We systematically explore the effectiveness of these features and their combinations, and most of the features such as BERT sentence embedding are used in $N$-best list rescoring for the first time and shown to be quite promising effect.

\begin{figure*}[t!]
    \centering
    \includegraphics[width=0.78\textwidth]{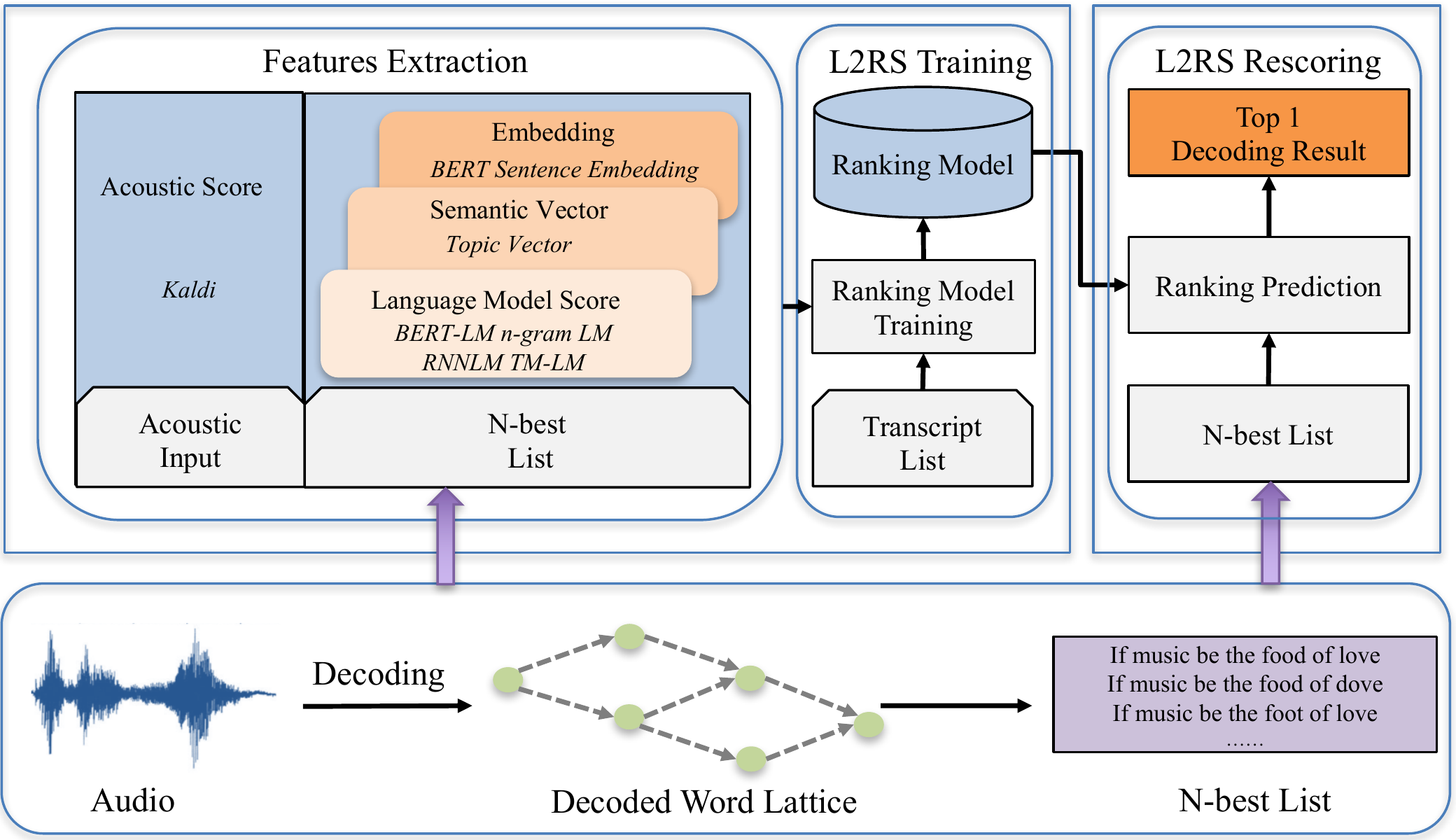}
    \caption{Flowchart of the proposed L2RS for ASR}
    \label{fig:ltr}
    \vspace{-10pt}
\end{figure*}

$\bullet$  We conduct extensive experiments based on a public dataset and experimental results shows that L2RS outperforms not only traditional rescoring methods but also its deep neural network counterparts such as RNNLM and EC-Model up to 20.67\% improvement which is quite substantial.

\section{Learning-to-Rescore}
\label{sec:models}

In this section, we first give the definition of the L2RS problem, followed by the description of the textual and acoustic features designed for L2RS. Finally, we describe the details of the rescoring model in L2RS.

\subsection{Problem Definition}
\label{sec:prob}

The pipeline of L2RS is listed in Fig.~\ref{fig:ltr}. Formally, the ASR system aims to find the optimal textual string $\mathbf{w^{*}}$ for a given acoustic input, denoted as $\mathbf{a}$, by the following equation:
\begin{equation}
\begin{aligned}
\mathbf{w^{*}} & = \arg \max (\log P_{LM}(\mathbf{w})  + \log P_{AM}(\mathbf{a|w}) + f(\phi(\mathbf{a}, \mathbf{w}))
\end{aligned}
\end{equation}
where $P_{LM}$ represents a back-off $n$-gram LM, $P_{AM}$ is an AM, $\phi(\mathbf{a}, \mathbf{w})$ is the feature-vector representation of pair $(\mathbf{a},\mathbf{w})$ including textual features as well as acoustic features, and $f(\cdot)$ is the rescoring function learned by L2RS approaches. The third component (i.e., $f(\phi(\cdot))$) is our contribution in this paper, which provides a new framework in ASR opening a lot of research opportunities.
During the decoding period, the ASR system generates the $N$-best list which is denoted as $\mathbf{w} = (\mathbf{w_1}, \cdots, \mathbf{w_j} ,\cdots, \mathbf{w_n})$, $j \in [1, n]$. The order list $(r_1, \cdots, r_n)$ of $N$-best hypotheses is decided based on the Word Error Rate (WER) of each hypothesis with the ground truth transcript.
This composes the training dataset $(\mathbf{x_i}, \mathbf{y_i})$, $i \in [1, m]$ used for L2RS, where $\mathbf{x_i} = (\phi(\mathbf{a_i}, \mathbf{w_{i,1}}), \cdots, \phi(\mathbf{a}, \mathbf{w_{i,n}}))$ and $\mathbf{y_{i}}=(r_{i,1}, \cdots, r_{i,n})$. During the L2RS prediction step, the ASR system generates the $N$-best list $\mathbf{w} = (\mathbf{w_1}, \cdots, \mathbf{w_n})$, and the final decoding result $\mathbf{w^{*}}$ can be obtained as follows:
\begin{equation}
\begin{aligned}
\mathbf{w^{*}} = \arg \max_{i \in [1, N]} f(\phi(\mathbf{a}, \mathbf{w_i}))
\end{aligned}
\end{equation}
\noindent L2RS learns $f(\phi(\mathbf{a}, \mathbf{w_i}))$ through a Learning-to-Rescore approach, which involves feature extraction, model training and rescoring.

\subsection{Textual Features}
\label{sec:text-feature}
The textual features used in L2RS are from the lexical level to the semantic level which belongs to six categories: $n$-gram LM, BERT Sentence Embedding, BERT LM, Probabilistic Topic Model LM, Topic Vector and RNNLM.

\noindent \textbf{$n$-gram LM} The $n$-gram LM is prominently used due to its simplicity and reliability. In L2RS, we use trigram LM trained using the transcript corpus with the SRILM\footnote{http://www.speech.sri.com/projects/srilm/} toolkit.

\begin{figure}[t!]
    \centering
    \includegraphics[width=0.30\textwidth]{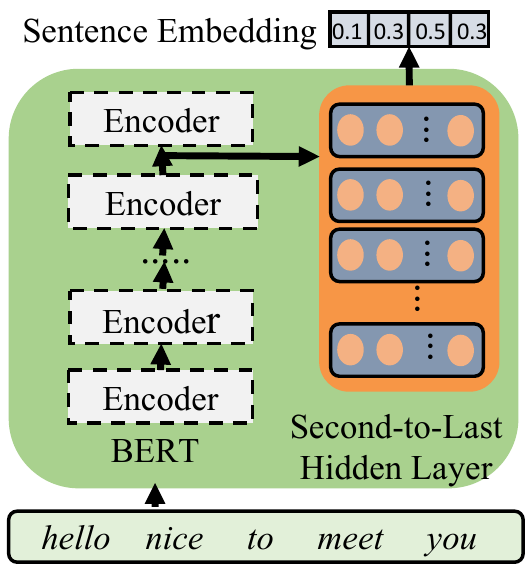}
    \caption{BERT Sentence Embedding}
    \label{fig:bwe}
\vspace{-10pt}
\end{figure}

\noindent \textbf{BERT Sentence Embedding}
BERT, or Bidirectional Encoder Representations from Transformers \cite{devlin2019bert}, is a powerful new language representation model proposed by Google and obtains the state-of-the-art results on various NLP tasks. The goal of BERT sentence embedding is to represent a variable length $N$-best hypothesis into a fixed length vector, e.g. ``\emph{hello, nice to meet you}'' to $[0.1, 0.3, 0.5, 0.3]$ shown in Fig.~\ref{fig:bwe}. Each element of this vector represents the semantics of the original sentence and this vector are further used in L2RS as a representation for each $N$-best hypothesis.

\noindent \textbf{BERT LM}
BERT can also be used as a LM \cite{wang2019bert} to evaluate the quality of the $N$-best hypotheses from the linguistic perspective. In L2RS, we use the perplexity given by a fine-tuned BERT model as a feature of the $N$-best hypotheses.

\noindent \textbf{{Probabilistic Topic Model LM}} Topic Modeling such as LDA \cite{blei2003latent} and SentenceLDA \cite{balikas2016topic} has the ability of capturing the semantic coherence of the $N$-best hypotheses. We first train a topic model based on the transcript corpus, which produces the topic-word distribution $\varphi_{kw}$ ($k \in [1, K]$ is the topic index and $w \in [1, W]$ is the word index). Next, we use the trained model to obtain the topic mixing proportion vector $\mathbf{\theta}_{d}$ of each hypothesis $d$, which represents the semantic meaning of this hypothesis. Based on these two parameters, we compute a transcript-specific unigram LM by:
\begin{equation}
\begin{aligned}
p(\mathbf{w}|\theta_{d}) = \sum_{k \in K} \theta_{dk}\varphi_{kw}
\end{aligned}
\end{equation}

\noindent \textbf{{Topic Vector}} Similar to Topic Model LM, L2RS directly uses the trained topic model to infer the $N$-best candidate's topic mixing proportion vector $\mathbf{\theta}$ and this vector is used as a topic representation for each $N$-best hypothesis.

\noindent \textbf{Neural Network-based LM} Neural network-based LMs are proven to be effective for $N$-best list rescoring in ASR systems.
We train a RNNLM \cite{mikolov2010recurrent} with the transcript corpus, and the perplexity of each hypothesis given by the RNNLM acts as a feature reflecting the quality of the hypothesis.

\subsection{Acoustic Feature}
The acoustic feature used in L2RS is the acoustic score given by the acoustic model. Specifically, in L2RS, we trained a ``chain'' model based on the training data using the Kaldi\footnote{https://github.com/kaldi-asr/kaldi} toolkit. It should be noted that other features such as speech embedding produced by Speech2Vec \cite{chung2018speech2vec} can also be used.

\subsection{Rescoring Model}
Learning to Rank \cite{liu2009learning} is a central problem for information retrieval. There are three categories for Learning to Rank: Pointwise approaches such as McRank \cite{li2008mcrank}, Pairwise approaches such as RankSVM \cite{herbrich2000large}, and Listwise approaches such as SVM MAP \cite{yue2007support}.
In L2RS, we choose RankSVM to train a rescoring model, and the learning of RankSVM is formalized as the following quadratic programming problem:
\begin{equation}
\begin{aligned}
\min_{w, \mathbf{\varepsilon} } \frac{1}{2}\left \| w \right \|^2 + C \sum_{i=1}^{m} \varepsilon_i \\
\textup{s.t. } y_i < w, \mathbf{\phi(\mathbf{a}, \mathbf{w}_{i})}- \mathbf{\phi(\mathbf{a}, \mathbf{w}_{j})} > \geq 1- \varepsilon_{i} \\
 \varepsilon_{i} \geq 0, i = 1, \cdots, m
\end{aligned}
\end{equation}
where $\mathbf{w}_{i}$ and $\mathbf{w}_{j}$ are two instances from the same $N$-best list, $\left \| \cdot  \right \|$ denotes $L_2$ norm, $m$ denotes the number of training instances, and $C > 0$ is a coefficient.

\section{Experiments}
\label{sec:experiments}
In this section we conduct experiments on a public dataset to verify the effectiveness of the proposed model.
\vspace{-10pt}
\subsection{Experiment Setup}
\label{sec:setup}

We use the public TED-LIUM dataset\footnote{https://www.openslr.org/51/} \cite{hernandez2018ted} in our experiment with the statistics listed in Table~\ref{tab:dataset}. 
For RankSVM\footnote{https://www.cs.cornell.edu/people/tj/svm\_light/svm\_rank.html}, the parameters $C$ is set to 10. For BERT, we first set up a pretrained BERT model\footnote{https://github.com/google-research/bert}, and then conduct fine-tuning using the transcript corpus of the training dataset. The dimension of BERT sentence embedding is set to 1024 using method similar as bert-as-service toolkit \cite{xiao2018bertservice}. For topic modeling, we use LightLDA \cite{yuan2015lightlda} and the number of the topics is set to 50. Following \cite{Liu2014Efficient}, we obtain 50-best for each utterances in the dataset.
Our algorithm is compared with the following baseline methods for $N$-best list rescoring: $n$-gram LM, RNNLM \cite{li2018recurrent}, BERT LM \cite{xiao2018bertservice}, Trigger-based DLM \cite{singh2007trigger}, Cache Model \cite{li2018recurrent}, EC-Model \cite{ogawa2018rescoring} and Neural Speech-to-Text LM (NS2TLM) \cite{tanaka2018neural}.
All experiments were conducted on a server with 314GB memory, 72 Intel Core Processor (Xeon), Tesla K80 GPU and CentOS.

\vspace{-5pt}
\begin{table}[ht!]
\caption{The statistics of the TED-LIUM dataset}
\centering
  \begin{tabular}{c|c|c|c}
    \toprule
    \textbf{} & \textbf{Train} & \textbf{Dev} & \textbf{Test} \\
	  \midrule
	  No. of transcripts & 774 & 8 & 11  \\
	  No. of words & 1.5M & 17.8k & 27.5k  \\
	  No. of segments & 56.8k & 0.6k & 1.5k  \\
	  Length of waves & 118hours & 1.72hours & 3.07hours \\
	  Frequency & \multicolumn{3}{c}{16kHz}  \\
      Language & \multicolumn{3}{c}{English}  \\
    \bottomrule
  \end{tabular}
  \label{tab:dataset}
  \vspace{-8pt}
\end{table}
%\vspace{-10pt}

\subsection{Experimental Results}
\label{sec:result}

\subsubsection{Normalized Discount Cumulative Gain (NDCG)}
\label{sec:ndcg}

Table \ref{tb:map} lists the rescoring performance of L2RS in terms of NDCG@$N$ \cite{manning2010introduction}. NDCG@$N$ is a measure widely used for reflecting the top $N$ quality of the ranking list, and the higher the better.
In most cases, the ASR system finally delivers the 1-best result from the rescored $N$-best list.
However, some tasks such as ASR in noisy environments or casual-style speech require multiple recognition hypotheses \cite{ogawa2018rescoring,mangu2000finding}.
From the result, we can see that compared with other methods, L2RS can produce better ranking list, which means not only the top 1 result is improved but also the whole ranking list is correctly ordered. Specifically, BERT sentence embedding is quite effective for L2RS and it has 14.58\% relative improvement over the baseline AM+$n$-gram LM rescoring method.
By incorporating all these features, L2RS(opt) achieves up to 20.67\% relative improvement over AM+$n$-gram baseline.

\begin{table}[ht!]
\vspace{-5pt}
\caption{NDCG@10 of Different $N$-best Rescoring Methods}
\centering
\begin{tabular}{l|c|c}
  \toprule
  \textbf{Model} & \textbf{Dev} & \textbf{Test}  \\
  \midrule
  AM + $n$-gram LM  & 0.5931 & 0.5859  \\
  L2RS(AM+$n$-gram LM)  & 0.5531  & 0.5082 \\
  L2RS(AM+RNNLM) &  0.6400 & 0.6108  \\
  L2RS(AM+BERT-LM) &  0.5148 & 0.5360  \\
  L2RS(AM+TM-LM) & 0.5333 & 0.5340  \\
  L2RS(BERT-WE)  & 0.7181 & 0.6713  \\
  L2RS(Topic-Vec)  & 0.4864 & 0.4760  \\
  L2RS(opt) & \textbf{0.7430} & \textbf{0.7070}  \\
  \bottomrule
\end{tabular}
\label{tb:map}
\vspace{-10pt}
\end{table}

\vspace{-10pt}

\subsubsection{Word Error Rate (WER)}
\label{sec:wer}
Since our ultimate goal is to improve ASR, we finally examine the effectiveness of L2RS method in terms of WER with results listed in Table~\ref{tb:wer}.
The ``Oracle'' WER is computed using the best result each time from the $N$-best list by comparing with the ground truth transcript, and it is the theoretical ceiling performance of all the rescoring methods. Among all these methods, RNNLM, BERT-LM, Trigger-based DLM, Cache Model, EC-Model, NS2TLM and L2RS(opt) have respectively 1.728\%, -0.083\%, -1.036\%, 0.026\%,
0.204\%, 0.506\% and 2.448\% improvement over the baseline $n$-gram LM method in Test dataset.
L2RS shows performance improvement over the state-of-the-art rescoring methods by a significant margin.
The experiment result validates that by incorporating more valuable features from the state-of-the-art NLP models, L2RS can benefit the current ASR system.

\begin{table}[ht!]
\caption{WER of Different $N$-best Rescoring Methods}
\centering
\begin{tabular}{l|c|c}
  \toprule
  \textbf{Model} & \textbf{Dev} & \textbf{Test}  \\
  \midrule
  $n$-gram LM & 21.999\% & 27.084\%  \\
  RNNLM  & 20.435\% & 25.356\% \\
  BERT-LM  & 22.302\% & 27.167\% \\
  Trigger-based DLM & 23.303\% & 28.120\%  \\
  Cache Model & 21.925\% & 27.058\%  \\
  EC-Model & 21.706\% & 26.880\%  \\
  NS2TLM & 21.200\% & 26.578\% \\
  \midrule
  L2RS(AM+$n$-gram LM) & 22.881\% & 28.778\%  \\
  L2RS(AM+RNNLM) & 21.391\% & 26.662\%  \\
  L2RS(AM+BERT-LM) & 23.359\% & 28.098\%  \\
  L2RS(AM+TM-LM) & 23.213\% & 28.109\%  \\
  L2RS(BERT-WE) & 20.250\% & 25.640\%  \\
  L2RS(Topic-Vec)  & 23.697\% & 29.658\% \\
  L2RS(opt) & \textbf{19.924\%} & \textbf{24.636\%}  \\
  \midrule
  Oracle & 16.538\% & 19.196\%  \\
  \bottomrule
\end{tabular}
\label{tb:wer}
\vspace{-15pt}
\end{table}

\vspace{-10pt}
\subsubsection{Quantitative Analysis of Features}
\label{sec:feature_quality}
\begin{figure}[h!]
\vspace{-0pt}
\centering
\includegraphics[width=0.40\textwidth]{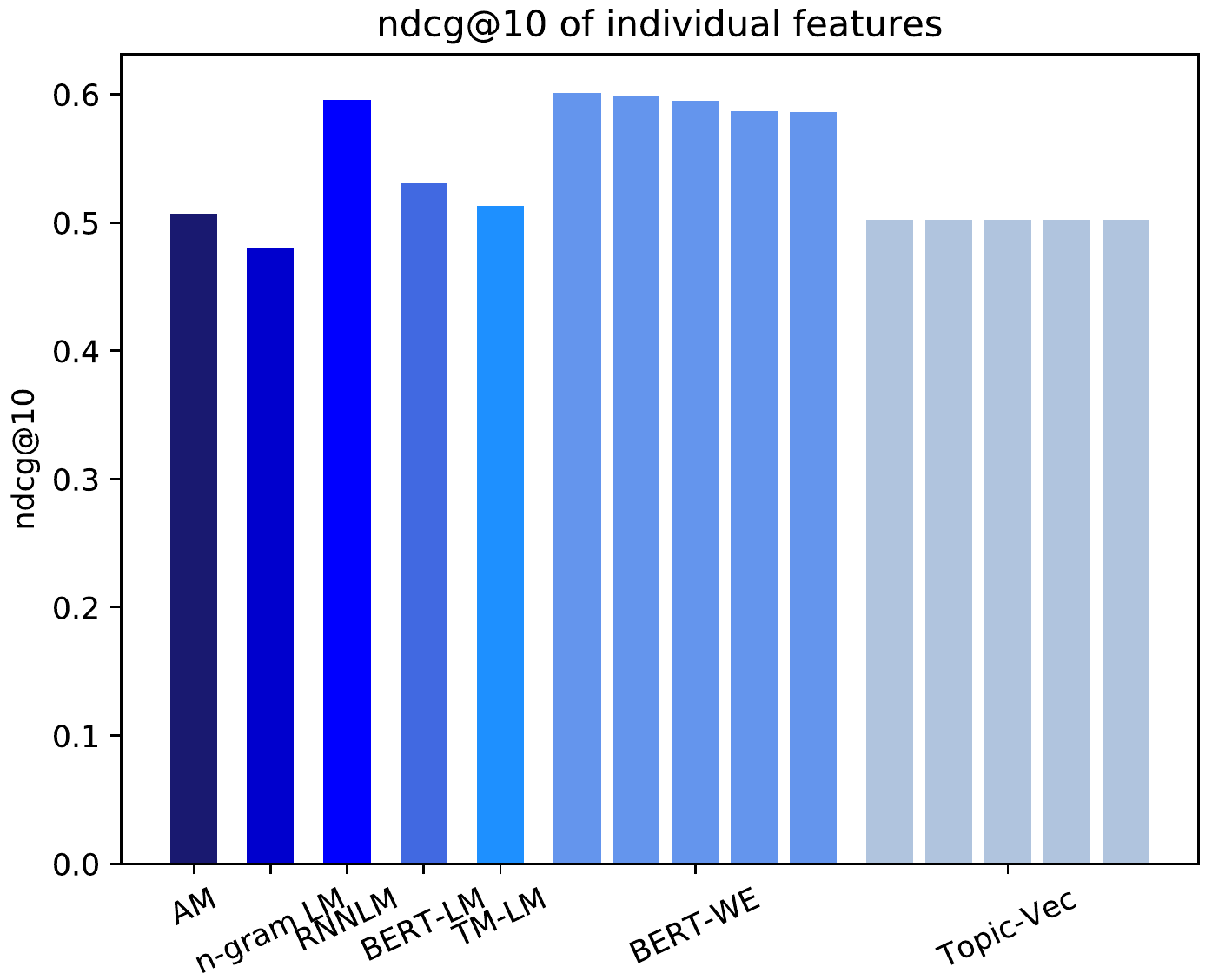}
\caption{Quality (NDCG@10) of Individual Features}
\label{fig:feature}
\vspace{-20pt}
\end{figure}

We use each dimension of these features to train a L2RS model and take the NDCG$@$10 as a measure to reflect the quality of the feature \cite{geng2007feature}.
The result is listed in Fig.~\ref{fig:feature} with the $x$-axis representing the feature category and the $y$-axis representing their NDCG values.
We can see that besides traditional AM and LM scores, other features also provide valuable information from different linguistic and semantic perspectives.
Features such as BERT sentence embedding, which is hard to be used under traditional rescoring pipeline, are even more effective than the RNNLM score. L2RS provides a flexible mechanism to explore the effects of these \emph{embedding} features and their combinations for $N$-best list rescoring.

\section{Conclusion}
\label{sec:conclusion}

In this paper, we propose a novel Learning-to-Rescore mechanism for ASR. L2RS formalizes the $N$-best list rescoring as a learning problem,
and incorporates comprehensive features with automatic optimized weights to form a rescoring model.
Experimental results have indicated that L2RS is quite effective for $N$-best list rescoring and opens a new door for ASR.
For future work, we will design neural L2RS models dedicated for ASR systems.

% References should be produced using the bibtex program from suitable
% BiBTeX files (here: strings, refs, manuals). The IEEEbib.bst bibliography
% style file from IEEE produces unsorted bibliography list.
% -------------------------------------------------------------------------
\footnotesize
\bibliographystyle{IEEEbib}
\bibliography{ltr}

\end{document}